\documentclass[10pt,twocolumn,letterpaper]{article}

\usepackage[pagenumbers]{cvpr} 

\definecolor{cvprblue}{rgb}{0.21,0.49,0.74}
\usepackage[pagebackref,breaklinks,colorlinks,allcolors=cvprblue]{hyperref}
\usepackage{multirow}
\usepackage{booktabs}
\usepackage{afterpage}
\usepackage{float}
\usepackage[table,xcdraw]{xcolor}
\usepackage[normalem]{ulem}
\useunder{\uline}{\ul}{}
\usepackage{subcaption}
\usepackage[accsupp]{axessibility}

\def\paperID{17922} 
\def\confName{CVPR}
\def\confYear{2026}

\title{CineMatte: Background Matting for Virtual Production and Beyond}

\author{
	Yuanjian He\footnotemark[1],
	Chen Zhang,
	Fasheng Chen,
	Jiangbo Cao \\
	Online Video Business Unit, Tencent PCG \\
	Shenzhen, China \\
	{\tt\small thomasyjhe@tencent.com, ichenzhang@tencent.com,
		fashengchen@tencent.com, brodycao@tencent.com}
}

\begin{document}

	\renewcommand{\thefootnote}{\fnsymbol{footnote}}

	\twocolumn[{%
		\renewcommand\twocolumn[1][]{#1}%
		\maketitle
		\includegraphics[width=1\linewidth]{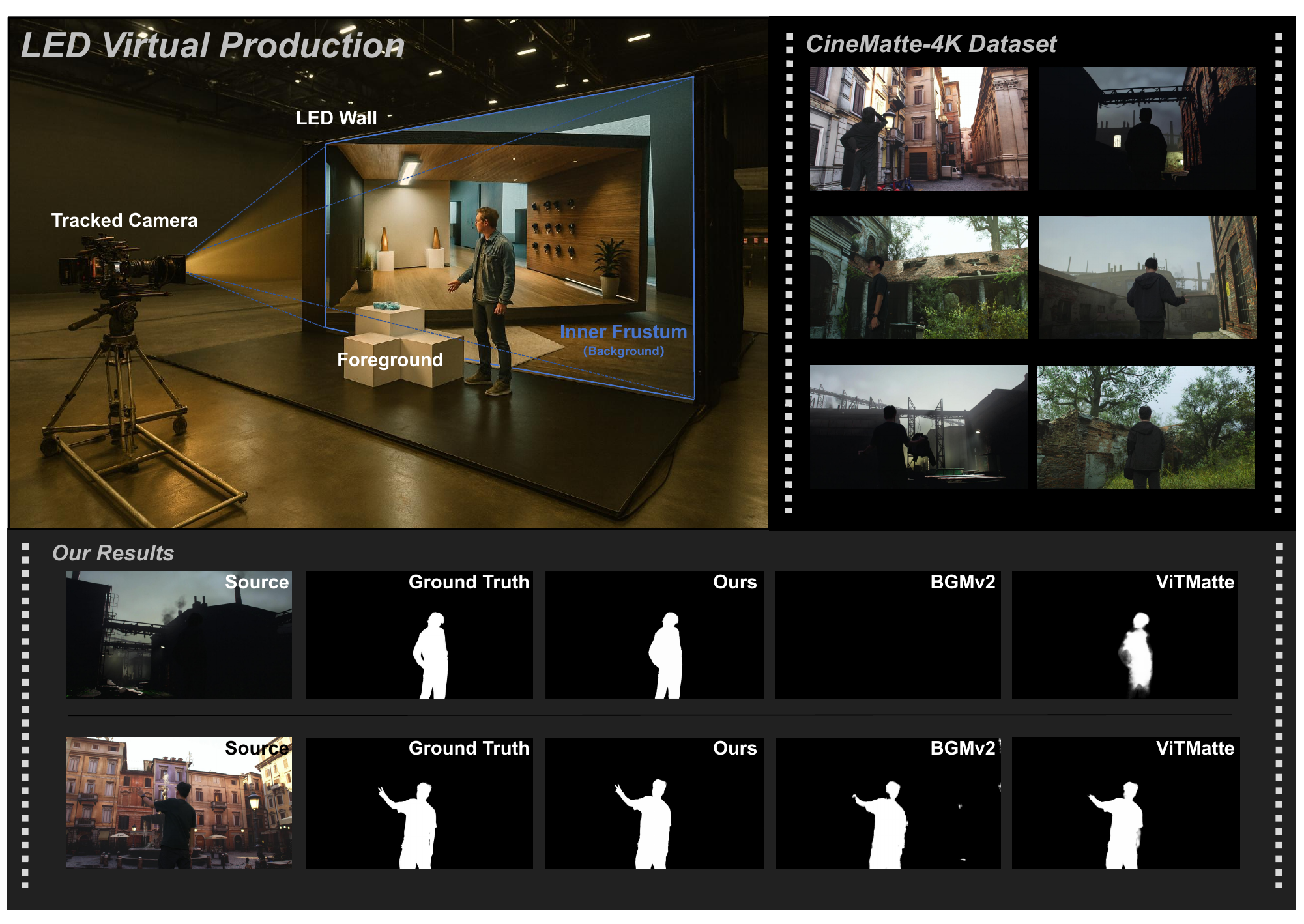}
		\captionof{figure}{We propose CineMatte, a background matting method for virtual production and beyond. }
		\vspace{1cm}
		\label{fig:teaser}
	}]

	\footnotetext[1]{Corresponding author.}
	\renewcommand{\thefootnote}{\arabic{footnote}}

	\begin{abstract}
	LED Virtual Production (VP) uses large LED volumes to render backgrounds in real time, enabling in‑camera visual effects but making post‑shot changes labor‑intensive. We address this with CineMatte, a robust background matting framework for VP and beyond. CineMatte employs a cross-attention‑conditioned design. Instead of concatenating the background with the input, CineMatte employs a Siamese, frozen DINOv3 Vision Transformer with shared weights to encode the input frame and the captured background separately. A cross‑attention module compares the two streams to predict the foreground, preserving pretrained semantics and improving robustness to background shifts. Previous ViT‑based matting models use a parallel convolutional “detail branch” to recover fine details, which can cause boundary artifacts in real‑world samples due to semantic misalignment with the backbone. We instead replace it with a pretrained, image‑guided feature upsampler, which largely mitigates the problem. We also collect CineMatte‑4K, a 4K HDR image–video dataset captured on a professional LED VP stage. To the best of our knowledge, the image subset is the first dataset for VP matting and is non‑synthetic, obtained via green‑screen insertion; the video subset includes camera motion with tracked trajectories so that arbitrary backgrounds can be rendered later with correct parallax. Across CineMatte‑4K and public benchmarks (VideoMatte240K, YouTubeMatte), CineMatte not only excels in VP but also generalizes robustly to real‑world footage.
\end{abstract}

	\section{Introduction}
LED Virtual Production (VP) replaces the traditional green screen with a wall of high‑brightness LED panels~\cite{pires2022survey, swords2024emergence}. As shown in Fig.~\ref{fig:teaser}, actors and physical props perform in front of this wall (often a wrap‑around “LED volume”). The production camera is tracked in 3D, and a game engine renders the virtual world from that exact viewpoint. The portion of the wall seen by the camera (the “inner frustum”) shows a high‑resolution, perspective‑correct image that updates every frame, so as the camera pans or dollies, parallax and depth cues look natural. Because the LEDs are bright, they also light the actors, producing reflections and colors that match the background and avoiding green‑screen spill. This workflow allows directors and actors to visualize the final pixel when filming on set, instead of imagining with a green screen. It also reduces post-production and works well with glass, hair, and other semi‑transparent or reflective materials that are hard to key on green. However, VP also makes it harder to revise a shot later: once a take is captured “in-camera,” changing or enhancing the background or tweaking lighting in post often requires heavy manual work. A robust background matting method would restore much of that flexibility for color matching, relighting, or background edits while preserving the on‑set benefits of VP.

Beyond virtual production, background matting~\cite{sengupta2020background, lin2020bgmv2} offers practical advantages over trimap-based~\cite{xu2017dim, yao2024vitmatte, chen2013knn} or segmentation-based approaches~\cite{yang2025matanyone, yu2021mgmatting} in certain applications. Capturing a clean background plate is often more feasible than labeling per-frame trimaps for complex, dynamic foregrounds. However, the trade-off is that a background is a slightly weaker semantic cue than a trimap: instead of being told foreground/unknown/background explicitly, the model must compare the input and the background, reason about scene changes, and be robust to viewpoint or content shifts.

\begin{figure*}[!t]
	\centering
	\includegraphics[width=1\linewidth]{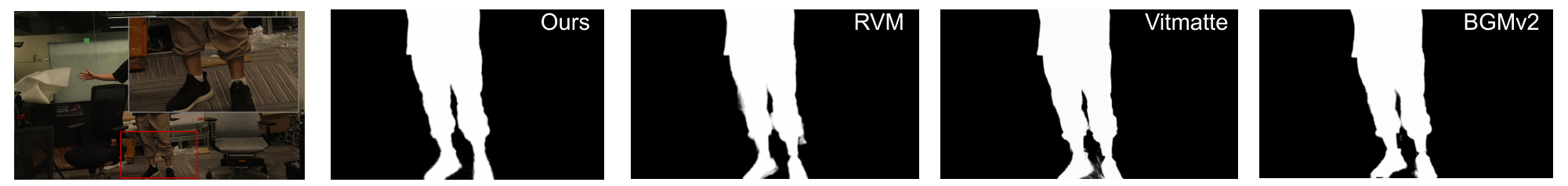}

	\caption{Real-world results: our method yields crisp boundaries and the most complete human matte, while baselines misclassify the socks as background.}
	\label{fig:real}
\end{figure*}

Prior background‑matting systems~\cite{sengupta2020background, lin2020bgmv2} are incapable of such highly demanding applications for several reasons: First, they are vulnerable to large background shifts, which are common in VP due to render‑to‑screen latency, imperfect hand‑eye calibration, and lighting variation. Second, their backbones are typically convolutional (e.g., ResNet~\cite{he2016resnet}): strong on edges and textures but weaker in semantic understanding than modern, self-supervised Vision Transformers (ViT)~\cite{dosovitskiy2020vit, caron2021emerging, oquab2023dinov2, simeoni2025dinov3, vaswani2017attention}. In cinematic settings, semantic robustness and completeness are critical. ViT‑based features (e.g., DINOv3~\cite{simeoni2025dinov3}) provide strong semantics, but attention tends to smooth high‑frequency detail~\cite{park2022vision}. To compensate, some transformer‑based methods~\cite{yao2024vitmatte, park2022matteformer} add a parallel convolutional branch to bring back edges and textures. In practice, this extra detail stream is not aligned with the backbone’s semantic features and can overfit to small matting datasets, creating artifacts in challenging real scenes (see Fig.~\ref{fig:real} for example). For image matting in general, a broader issue is generalization: many methods perform impressively on synthetic benchmarks yet struggle in the wild. If the backbone is already strong, minimally adapting it to the task can help preserve the generalization learned from billions of images.

Motivated by these observations, we propose \textbf{CineMatte}, a background matting method designed for VP and beyond. We use a \textbf{frozen} DINOv3 ViT as our encoder to keep its pretrained semantics intact. Instead of concatenating the condition (background) as extra channels and fine-tuning the whole encoder, we employ a cross-attention-conditioned design with a frozen, weight-sharing encoder: one processes the input frame and the other processes the background. The image stream is conditioned on the background via a cross-attention-based module between the two streams, allowing the network to compare content without disrupting the encoder’s pretraining distribution. Cross-attention conditioning improves robustness to background shifts, and freezing the encoder reduces overfitting, yielding stronger generalization to in-the-wild footage. To recover high-frequency detail, we attach a pretrained feature upsampler that reconstructs crisp boundaries while staying aligned with the ViT’s semantic features. Despite the frozen encoder, CineMatte matches strong baselines on synthetic datasets and remains stable and artifact-resistant in difficult real scenes.

On the other hand, data remains a bottleneck: most image matting datasets are synthetic, most video datasets assume static cameras, and, to our knowledge, prior work lacks a dedicated virtual production matting dataset.\textbf{} We collect $\textbf{CineMatte‑4K}$, a 4K HDR image–video dataset captured on a professional VP stage. For images, we use a green‑screen insertion workflow to obtain high‑quality ground truth: during capture, the LED wall alternates between the target scene and a green screen; the scene frame serves as the input, and the ground‑truth alpha is obtained by manually matting the corresponding green‑screen frame. This yields a non‑synthetic dataset for background matting and virtual production. For videos, we record tracked camera trajectories together with green-screen foregrounds, enabling later rendering of arbitrary backgrounds with correct parallax. This enables dynamic footage with real camera motion and complements existing static, synthetic video datasets.

In summary, our main contributions are as follows:

• We introduce CineMatte, a cross-attention-conditioned background matting framework that keeps a frozen, weight-shared DINOv3 ViT for the input frame and the captured background, and conditions the image stream via a Foreground–Background Alignment Module (FBAM). This preserves pretrained semantics and improves robustness to background shifts.

• We adopt a pretrained, image-guided feature upsampler (JAFAR-style~\cite{couairon2025jafar}) to recover crisp boundaries while staying semantically aligned with the ViT. We further explore a window-attention variant that reduces training VRAM by $\approx$ 50\% with minimal performance drop.

• We collect CineMatte‑4K, a 4K HDR image–video dataset captured on a professional virtual production stage. The image subset is non-synthetic with mattes obtained from green screen insertion; the video subset contains tracked camera trajectories for parallax-correct rendering.

• Across VideoMatte240K~\cite{lin2020bgmv2}, YouTubeMatte~\cite{yang2025matanyone}, and our VP dataset CineMatte‑4K, CineMatte delivers competitive results on MAD/MSE/Grad/Conn and dtSSD, and demonstrates robustness under synthetic background‑shift stress tests.
	\begin{figure}[t]
	\centering
	\includegraphics[width=1\linewidth]{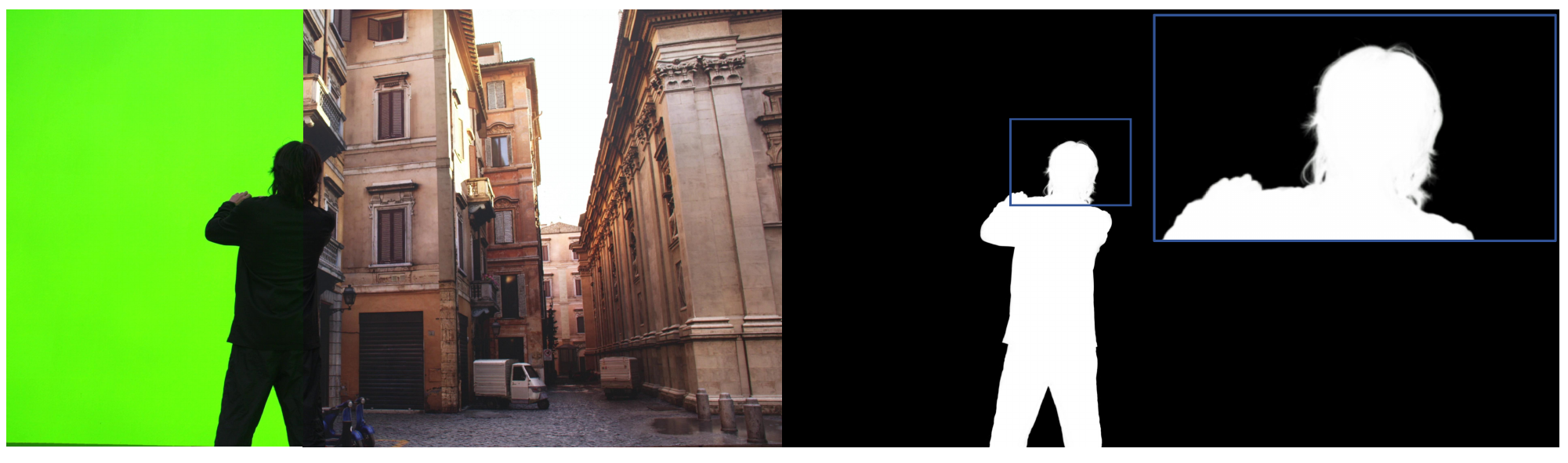}

	\caption{Creation of the CineMatte-4K dataset}
	\label{fig:dataset}
	\vspace{-0.3cm}
\end{figure}

\begin{figure*}[t]
	\vspace{-0.5cm}
	\centering
	\includegraphics[width=0.95\linewidth]{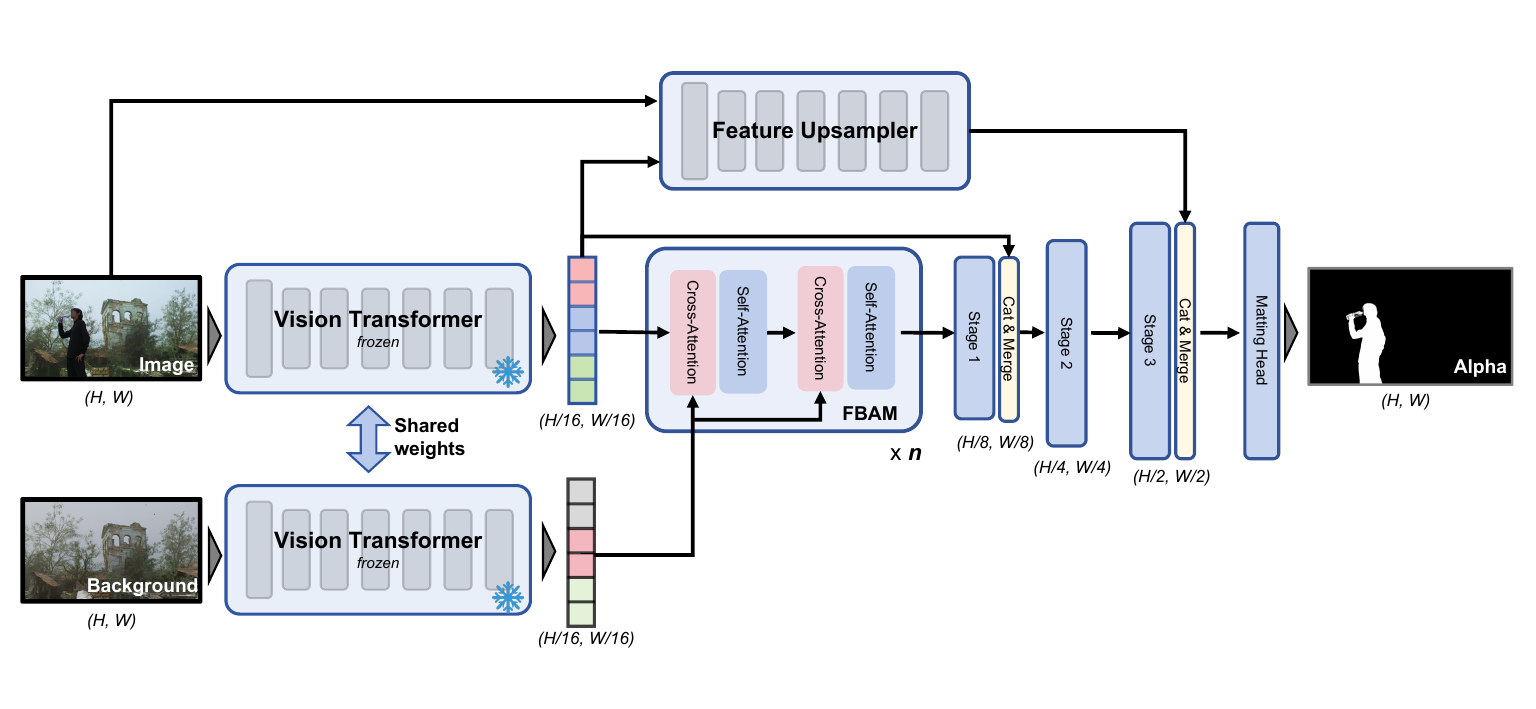}
	\vspace{-0.8cm}
	\caption{Overview of CineMatte. A frozen Siamese DINOv3~\cite{simeoni2025dinov3} encoder extracts tokens from the input image and background. FBAM performs cross-attention between the two streams, and a decoder progressively upsamples features. A JAFAR-style~\cite{couairon2025jafar} feature upsampler injects high-resolution boundary details before alpha prediction.}
	\label{fig:overall_framework}
	\vspace{-0.3cm}
\end{figure*}

\section{Related work}
{
	\setlength{\parindent}{0cm}
	\textbf{Alpha Matting.}
}
Early natural image matting methods formulated alpha estimation under carefully chosen priors and user input (e.g., trimaps)~\cite{ruzon2000alpha, wang2005unified, wang2007optimized, rhemann2008highres, rhemann2010psf, he2011globalsampling, lee2011nonlocal, gastal2010shared}. Closed-Form Matting~\cite{levin2008closedform} derived a sparse quadratic objective in $\alpha$ from local smoothness assumptions; and KNN Matting~\cite{chen2013knn} leveraged nonlocal neighbors to obtain a closed-form solution robust to sparse user marks.

With deep learning, convolutional image matting~\cite{xu2017dim, yu2021mgmatting, lu2019indexnet, li2020gca, dai2021a2u, li2020gfm, li2021p3m, park2022matteformer} and video matting~\cite{ke2022modnet, lin2022rvm, yang2025matanyone, huynh2024maggie, huang2023end} advanced rapidly. Deep Image Matting (DIM)~\cite{xu2017dim} introduced an encoder–decoder with a refinement head, setting a strong baseline for learning-based alpha prediction. Beyond trimaps, mask/segmentation-conditioned matting reduces annotation burden: MG Matting~\cite{yu2021mgmatting} accepts coarse masks and progressively refines uncertain regions. More recently, transformer-based matting methods have started to gain attention~\cite{cai2022transmatting, yao2024vitmatte, luthra2021eformer, li2024vmformer, park2022matteformer}. ViTMatte~\cite{yao2024vitmatte} shows that plain ViT backbones with a lightweight detail module can yield state-of-the-art image matting from trimaps.

Background matting offers an alternative to conditional matting by leveraging a captured clean background to guide matte estimation. This is convenient for video conferencing, imperfect green screens, and other cinematic applications. BGM~\cite{sengupta2020background} first demonstrated robust background‑conditioned matting from casual captures. BGMv2~\cite{lin2020bgmv2} achieved real‑time, high‑resolution performance by pairing a low‑resolution base network with patch‑wise high‑resolution refinement.

{
	\setlength{\parindent}{0cm}
	\textbf{Vision Foundation Models.}
}
ResNet~\cite{he2016resnet} established residual learning as the default convolutional backbone, striking a strong balance between efficiency and detail recovery for dense prediction~\cite{chen2017deeplabv3, zhao2017pspnet, lin2017refinenet, sun2019hrnet, xie2017resnext}. Vision Transformers (ViT)~\cite{dosovitskiy2020vit} replaced convolutions with global self-attention over patch tokens, bringing powerful long-range context. ViT has already been adapted to a variety of downstream applications~\cite{ranftl2021dpt, xie2021segformer, li2022vitdet, kirillov2023sam,weinzaepfel2022croco,weinzaepfel2023crocov2}. Self-supervised ViT pretraining has since produced general-purpose "foundation" encoders~\cite{caron2021emerging,oquab2023dinov2, simeoni2025dinov3}. DINOv3~\cite{simeoni2025dinov3} scales self-supervised ViTs and introduces Gram anchoring to maintain high-quality dense features, yielding strong transfer to pixel-level tasks without task-specific supervision. Complementary to strong encoders, high-fidelity reconstruction of spatial detail can be improved by feature upsamplers~\cite{couairon2025jafar, fu2024featup, suri2024lift, zhou2024refreshed}. JAFAR~\cite{couairon2025jafar} learns an attention-based, image-guided upsampler that lifts low-resolution backbone features to arbitrary resolutions while preserving semantic alignment, benefiting dense tasks that demand crisp, boundary-aligned signals.

\section{Methods}

The overall architecture is shown in Fig.~\ref{fig:overall_framework}.

{
	\setlength{\parindent}{0cm}
	\textbf{Design rationale.}
}
Early-fusing the background by channel concatenation forces the encoder to relearn its input statistics and tends to overfit small matting datasets. We instead keep the encoder frozen and inject background cues via token-level cross-attention, which (i) preserves pretrained semantics, (ii) is inherently more tolerant to background misalignment, and (iii) pairs naturally with an image-guided upsampler to restore high-frequency details without semantic misalignment with the backbone feature.

\subsection{Vision Transformer}
A Vision Transformer~\cite{dosovitskiy2020vit} models an image as a sequence of fixed-size patch tokens and processes this sequence with Transformer encoder blocks.

We denote the per-layer token sequence by $\mathbf{x}_{p}^{(l)} \in \mathbb{R}^{N \times D}$, where $p$ indicates the patchified token and $l$ is the transformer layer index. The input image $\mathbf{x} \in \mathbb{R}^{H \times W \times 3}$ is split into $N = HW/P^2$ non-overlapping $P \times P$ patches and projected to $\mathbf{x}_{p}^{(0)} \in \mathbb{R}^{N \times D}$ with positional encodings. For the $l$-th layer:

\begin{equation}
	\mathbf{x}^{(l) \prime}_{p} = \mathbf{x}_{p}^{(l)} + \mathrm{MHSA}\!\big(\mathrm{LN}(\mathbf{x}_{p}^{(l)})\big),
\end{equation}
\begin{equation}
	\mathbf{x}^{(l + 1)\prime}_{p} = \mathbf{x}^{(l) \prime}_{p} + \mathrm{MLP}\!\big(\mathrm{LN}(\mathbf{x}^{(l) \prime}_{p})\big).
	\label{eq:vit-mlp}
\end{equation}

After $L$ layers, we obtain $\mathbf{x}_{p}^{(L)} \in \mathbb{R}^{N \times D}$, which is fed to the subsequent modules. For clarity, bold symbols denote tensors with spatial or token dimensions, while regular symbols denote per-layer sequences. We use ViT-L/16, token grid is 1/16 of input.
\paragraph{DINOv3 essentials.}
DINOv3 is designed to produce strong \emph{dense} features without labels by training ViTs with self-supervision at scale. It combines instance discrimination, masked patch prediction, and uniformity regularization. To preserve feature locality during extended training, DINOv3 introduces Gram anchoring, which stabilizes pairwise patch similarities by constraining the student patch features $X_S$ to match those of a temporal teacher $X_T$:
\begin{equation}
	\mathcal{L}_{\mathrm{Gram}} \;=\; \big\|\,X_S X_S^\top \;-\; X_T X_T^\top \,\big\|_F^2.
\end{equation}

We freeze a DINOv3 encoder; this preserves robust semantics and high-quality token maps, which our FBAM and decoder adapt to background matting.
\begin{figure}[t]
	\centering
	\includegraphics[width=0.8\linewidth]{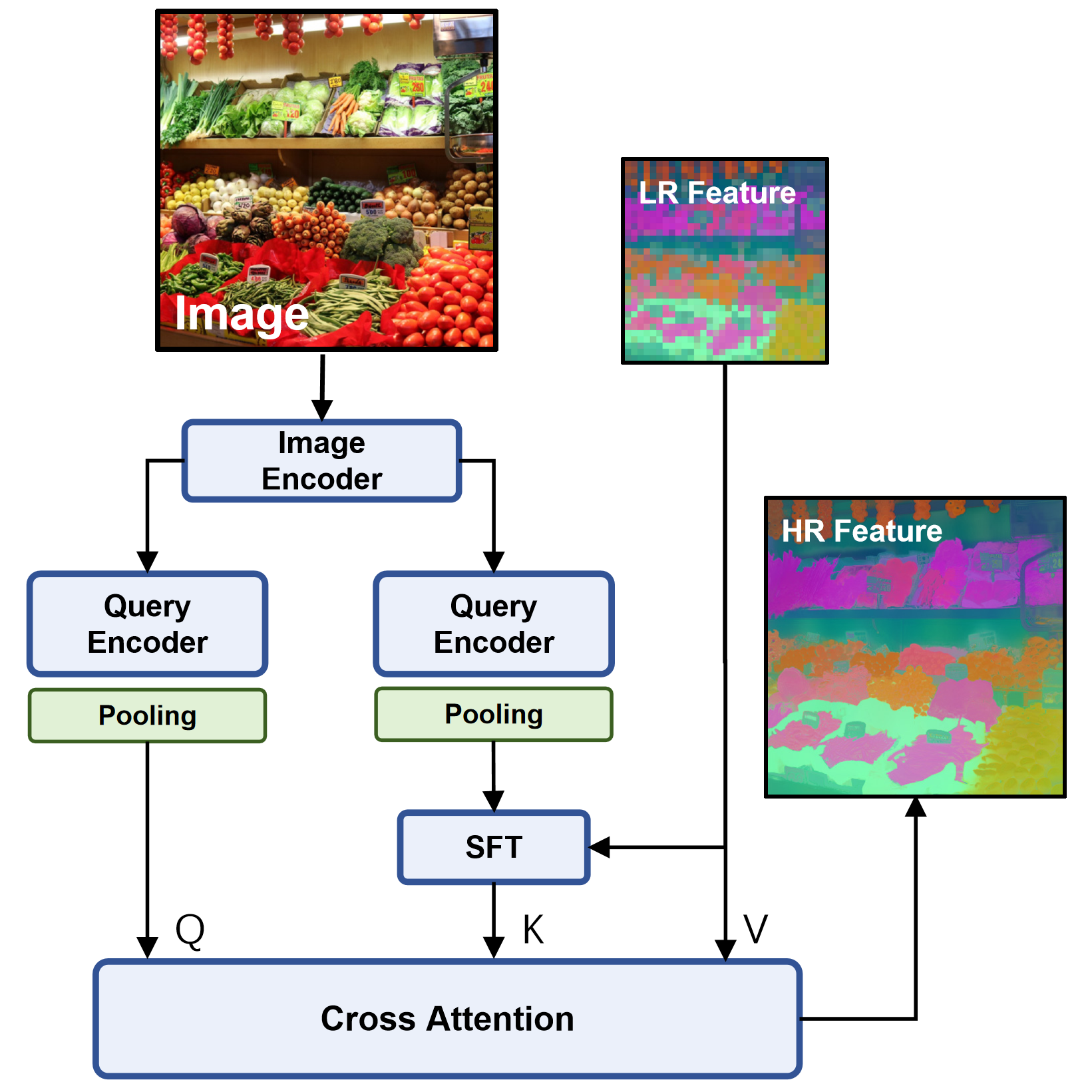}
	\caption{JAFAR-style feature upsampler. It recovers high-resolution details while staying aligned with backbone semantics.}
	\label{fig:feature_upsampler}
\end{figure}

\subsection{Foreground-Background Alignment Module}

Most conditional matting methods inject conditions (background, mask, trimap) via channel concatenation. This forces the encoder to adapt to an expanded input and typically requires end‑to‑end retraining, which can erode the knowledge in a strong backbone. As a result, such models often produce artifacts on real, out‑of‑distribution samples.

Concat‑based conditioning is also brittle to background shifts, which are common in LED virtual production. In practice, the rendered background on the LED wall can lag the camera feed by on the order of hundreds of milliseconds, and hand–eye calibration is often imperfect in working environments, leading to persistent misalignment. In this situation, prior background‑matting methods can degrade severely (see  Table~\ref{table_ablation_bg_shift}). We hypothesize this brittleness arises from early fusion via concatenation, the limited shift tolerance of convolutions (padding/stride and fixed, local kernels), and overfitting to small, synthetic datasets.

Instead, we keep the backbone frozen and inject the condition through our Foreground–Background Alignment Module (FBAM), which lets the image and background streams interact directly at the token level via cross-attention. This preserves pretrained semantics while improving robustness to background misregistration.

Given backbone image and background features $\mathbf{x}_i, \mathbf{x}_b \in \mathbb{R}^{N \times D}$, we apply a lightweight adaptation layer:
\begin{equation}
	\phi(\mathbf{x}) = \mathrm{ReLU}\!\big(\mathrm{GN}(\mathrm{Conv}_{1\times1}(\mathbf{x}))\big),
\end{equation}
\begin{equation}
	\mathbf{x}_i^{(0)} = \phi(\mathbf{x}_i), \quad \mathbf{x}_b^{(0)} = \phi(\mathbf{x}_b).
\end{equation}

We then stack $n$ FBAM layers to progressively refine the image features through
self-attention and cross-attention with the background. For the $l$-th layer ($l=1,\ldots,n$):
\begin{equation}
	\mathbf{x}^{(l)}_{sa} = \mathbf{x}^{(l-1)}_i + \mathrm{MHSA}\big(\mathrm{LN}(\mathbf{x}^{(l-1)}_i)\big),
\end{equation}
\begin{equation}
	\mathbf{x}^{(l)}_{ca} = \mathbf{x}^{(l)}_{sa} + \mathrm{MCA}\big(\mathrm{LN}_Q(\mathbf{x}^{(l)}_{sa}), \;\mathrm{LN}_{KV}(\mathbf{x}^{(0)}_b)\big),
\end{equation}
\begin{equation}
	\mathbf{x}^{(l)}_i = \mathbf{x}^{(l)}_{ca} + \mathrm{MLP}\big(\mathrm{LN}(\mathbf{x}^{(l)}_{ca})\big).
\end{equation}

MCA denotes multi‑head cross‑attention. The output $\mathbf{x}^{(n)} \in \mathbb{R}^{N \times D}$ is reshaped to a 2D feature map $\mathbf{x}_o \in \mathbb{R}^{h_b \times w_b \times D}$ for the spatial decoder, where $h_b \times w_b = N$.

FBAM injects the background condition without altering the backbone, enabling both streams to reason about foreground changes while preserving pretrained semantics and reducing boundary artifacts.

\begin{figure*}[t]
	\centering
	\includegraphics[width=0.8\linewidth]{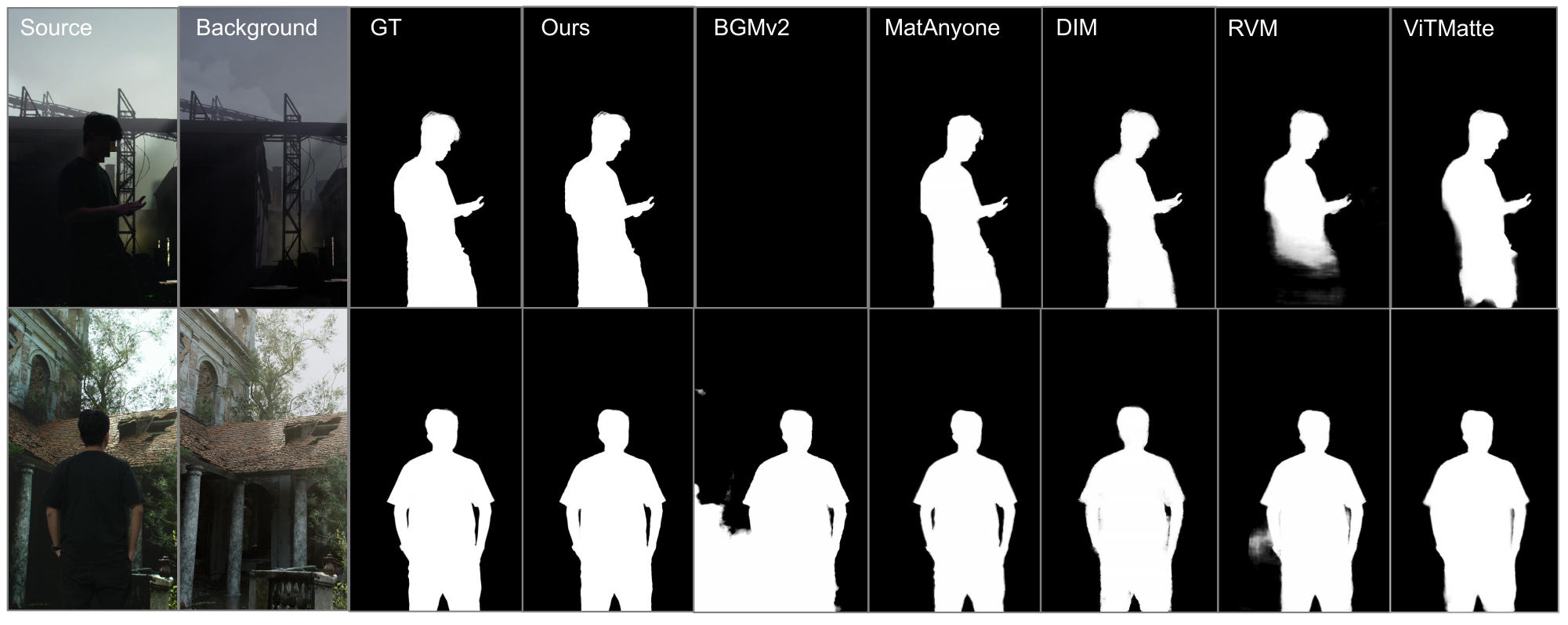}

	\caption{More samples on the CineMatte-4K dataset. Faces are blurred for anonymity.}
	\label{fig:more_samples}
	\vspace{-0.3cm}
\end{figure*}
\subsection{Feature Upsampler}
Although DINOv3 provides robust dense semantics, standard transformer backbones still tend to wash out high-frequency details (e.g., hair, thin edges) through patchification and attention~\cite{park2022vision}. Prior work (e.g., ViTMatte~\cite{yao2024vitmatte}) adds a parallel lightweight CNN branch to recover details, but the resulting high-frequency cues often misalign with backbone semantics, causing boundary artifacts.

To address this, we seek an architecture that preserves high‑resolution detail while remaining aligned with the backbone's semantics. We adopt JAFAR~\cite{couairon2025jafar} as our feature upsampler (Fig.~\ref{fig:feature_upsampler}).

We denote by $d$ the internal channel dimension of the upsampler's encoders and by $D$ the channel dimension of the backbone features. The source image first passes through an encoder to produce an intermediate feature map, $I_E \in \mathbb{R}^{H \times W \times d}$. Two lightweight encoders, followed by spatial pooling, generate queries and pre‑keys:
\begin{equation}
	Q = \mathrm{Pool}_{h \times w}\big(\mathrm{Enc}_q(I_E)\big),
\end{equation}
\begin{equation}
	\tilde{K} = \mathrm{Pool}_{h_b \times w_b}\big(\mathrm{Enc}_k(I_E)\big).
\end{equation}

$h \times w$ are the expected output resolution. Keys are then modulated with scale and shift parameters derived from the backbone feature:
\begin{equation}
	\gamma, \beta = \mathrm{Proj}(F_{\mathrm{lr}}),  \qquad K = \gamma \odot \tilde{K} + \beta.
\end{equation}
where $\odot$ denotes element‑wise multiplication. Cross‑attention is then computed between $Q \in \mathbb{R}^{h \times w \times d}$ and $K \in \mathbb{R}^{h_b \times w_b \times d}$, with values given by the backbone feature, $V = F_{\mathrm{lr}} \in \mathbb{R}^{h_b \times w_b \times D}$, which encourages the high‑resolution branch to stay aligned with the backbone's semantics. In our implementation we upsample the backbone feature by 8× per side.

While the original JAFAR employs global cross-attention, we further explore a variant computing attention within non-overlapping $m \times m$ windows to bound the quadratic cost in sequence length. This reduces training VRAM by approximately 50\% with a slight performance drop; all reported results in this paper use the full global attention variant.
\subsection{Decoder}
Our decoder adopts a DPT-style~\cite{ranftl2021dpt} progressive upsampling design. Given the FBAM output
$\mathbf{x}_o \in \mathbb{R}^{h_b \times w_b \times D}$, we apply 3
decoder stages. At stage $s$, we perform 2x bilinear upsampling to obtain $\mathbf{h}^{(s)}$ followed by feature
refinement via a residual unit:
\begin{equation}
	\mathbf{h}^{(s)'} = \mathrm{ReLU}\big(\mathrm{GN}(\mathrm{Conv}_1(\mathbf{h}^{(s)}))\big),
\end{equation}
\begin{equation}
	\mathbf{x}^{(s)} = \mathbf{h}^{(s)} + \mathrm{GN}\big(\mathrm{Conv}_2(\mathbf{h}^{(s)'})\big),
\end{equation}
where $\mathrm{Conv}_1$ and $\mathrm{Conv}_2$ are $3\times3$ convolutions. A $1\times1$ projection is applied to the skip connection when input/output channels differ.

At stage 1 ($1/8$ resolution), we incorporate the backbone output via a skip connection:
the backbone features are upsampled by $2\times$, adapted with a residual unit, and
merged with the main stream using a $1\times1$ convolution. At stage 3 ($1/2$ resolution),
high-resolution features from the feature upsampler are similarly adapted and merged.
The final decoder output is processed by a lightweight matting head with one additional
$2\times$ upsampling to produce the full-resolution alpha matte.

\section{CineMatte-4K dataset}
To our knowledge, prior work lacks a dedicated virtual production matting dataset. Virtual production lets us capture non-synthetic footage while still obtaining clean ground-truth alpha by temporarily inserting a green screen on the LED walls. Data were captured in a professional virtual production stage with over 300\,m$^2$ of LED walls. All shots are 4K HDR, recorded on a Sony CineAltaV 2 camera with an Angenieux Type EZ-2 lens. Ground-truth alpha mattes were created by human experts and quality-checked. The dataset includes an image subset with \textbf{500} samples (100 reserved for testing). The video dataset contains \textbf{30} clips, each around 10 seconds with a frame rate of 30 FPS.

\subsection{CineMatte-4K-Image}
As illustrated in Fig.~\ref{fig:dataset}, the actor remains still while the LED wall alternates between the target scene and a clean green screen. The camera records both conditions. We take the green-screen frames and produce alpha mattes via human annotation, then pair each matte with the corresponding scene frame. Every image--alpha pair is manually reviewed to ensure high matte quality and pixel-level alignment.

\begin{table*}[t]
	\centering
	\caption{Results and ablation, shown side-by-side.}
	\label{tab:results_ablation_side_sub}
	\begin{subtable}[t]{0.70\textwidth}
		\centering
		\small
		\caption{Results on VideoMatte240K~\cite{lin2020bgmv2} and YouTubeMatte~\cite{yang2025matanyone}. Best in bold; second underlined.}
		\label{result_public_dataset}
		{\setlength{\tabcolsep}{3.5pt}%
		\begin{tabular}{@{}llllllllll@{}}
			\toprule
			& \multicolumn{5}{c}{VideoMatte240k~\cite{lin2020bgmv2}}                                                                                                                                                                                              & \multicolumn{4}{c}{YouTubeMatte~\cite{yang2025matanyone}}                                                                                                                                  \\ \midrule
			\multicolumn{1}{l|}{Method}       & MAD$\downarrow$                                    & MSE$\downarrow$                                     & Grad$\downarrow$                                    & Conn$\downarrow$                                    & \multicolumn{1}{l|}{dtSSD$\downarrow$ }                                  & MAD$\downarrow$                                     & MSE$\downarrow$                                     & Grad$\downarrow$                                    & Conn$\downarrow$                                    \\ \midrule
			\multicolumn{1}{l|}{MODNet~\cite{ke2022modnet}}       & 6.732                                  & 1.937                                  & 7.523                                  & 24.050                                 & \multicolumn{1}{l|}{2.012}                                  & 18.806                                 & 16.713                                 & 15.935                                 & 129.569                                \\
			\multicolumn{1}{l|}{P3M~\cite{li2021p3m}}          & 6.430                                  & 2.425                                  & 6.949                                  & 20.421                                 & \multicolumn{1}{l|}{2.511}                                  & 24.087                                 & 21.727                                 & 17.016                                 & 179.835                                \\
			\multicolumn{1}{l|}{RVM~\cite{lin2022rvm}}          & 4.742                                  & 0.435                                  & 5.927    & \cellcolor[HTML]{E7E7E7}{\ul 6.424}    & \multicolumn{1}{l|}{1.503} & 15.481                                 & 9.633                                  & 14.764                                 & 70.615                                 \\
			\multicolumn{1}{l|}{MGM~\cite{yu2021mgmatting}}          & 7.641                                  & 1.661                                  & 9.219                                  & 13.474                                 & \multicolumn{1}{l|}{2.090}                                  & 20.470                                 & 13.894                                 & 12.556                                 & 115.834                                \\
			\multicolumn{1}{l|}{MatAnyone~\cite{yang2025matanyone}}    & 4.902                                  & 0.769                                  & \cellcolor[HTML]{E7E7E7}{\ul 5.907}                                 & 9.309                                 & \multicolumn{1}{l|}{\cellcolor[HTML]{C0C0C0}\textbf{1.427}}                                  & 2.667                                  & 1.017                                  & 10.942                                 & \cellcolor[HTML]{E7E7E7}{\ul 8.243}    \\
			\multicolumn{1}{l|}{KNN~\cite{chen2013knn}}          & 10.109                                 & 2.970                                  & 24.289                                 & 40.506                                 & \multicolumn{1}{l|}{4.745}                                  & 8.407                                  & 3.754                                  & 24.402                                 & 44.599                                 \\
			\multicolumn{1}{l|}{ViTMatte~\cite{yao2024vitmatte}}     & \cellcolor[HTML]{E7E7E7}{\ul 4.616}    & 0.447                                  & 6.921                                  & \cellcolor[HTML]{C0C0C0}\textbf{6.132} & \multicolumn{1}{l|}{1.605}                                  & \cellcolor[HTML]{E7E7E7}{\ul 1.810}    & \cellcolor[HTML]{E7E7E7}{\ul 0.483}    & \cellcolor[HTML]{E7E7E7}{\ul 7.562}    & \cellcolor[HTML]{C0C0C0}\textbf{8.152} \\
			\multicolumn{1}{l|}{DIM~\cite{xu2017dim}}          & 6.105                                  & 0.856                                  & 9.852                                  & 17.322                                 & \multicolumn{1}{l|}{2.120}                                  & 5.199                                  & 1.438                                  & 14.513                                 & 28.815                                 \\ \midrule
			\multicolumn{1}{l|}{BGMv2~\cite{lin2020bgmv2}}        & 5.202                                  & 0.854                                  & 7.059                                  & 11.198                                 & \multicolumn{1}{l|}{1.801}                                  & 2.142                                  & 0.648                                  & 8.369                                  & 12.800                                 \\
			\multicolumn{1}{l|}{\textbf{CineMatte}}                         & \cellcolor[HTML]{C0C0C0}\textbf{4.397} & \cellcolor[HTML]{C0C0C0}\textbf{0.341} & \cellcolor[HTML]{C0C0C0}\textbf{5.774} & 8.792                                  & \multicolumn{1}{l|}{\cellcolor[HTML]{E7E7E7}{\ul 1.492}}                         & \cellcolor[HTML]{C0C0C0}\textbf{1.779} & \cellcolor[HTML]{C0C0C0}\textbf{0.447} & \cellcolor[HTML]{C0C0C0}\textbf{7.267} & 12.277                                 \\ \bottomrule
		\end{tabular}
			}
	\end{subtable}
	\hfill
	\begin{subtable}[t]{0.28\textwidth}
		\centering
		\small
		\caption{Ablation on number of FBAM layers}
		\label{ablation_fbam}
		\begin{tabular}{@{}lll@{}}
			\toprule
			Num      & MAD$\downarrow$             & MSE$\downarrow$             \\ \midrule
			\multicolumn{1}{l|}{1} & 4.432          & 0.352          \\
			\multicolumn{1}{l|}{\textbf{2}} & \textbf{4.397} & \textbf{0.341} \\
			\multicolumn{1}{l|}{3} & 4.486          & 0.402          \\
			\multicolumn{1}{l|}{4} & 4.621          & 0.423          \\ \bottomrule
		\end{tabular}
	\end{subtable}
	\vspace{-0.2cm}
\end{table*}

\subsection{CineMatte-4K-Video}
Current video matting datasets~\cite{lin2020bgmv2, yang2025matanyone} are mostly synthetic: a 2D cutout foreground is composited onto a static background or a background video. This setup struggles to reproduce view-consistent geometry changes (e.g., parallax and perspective) that naturally arise with camera motion. Since real-world shots almost always involve a moving camera and noticeable viewpoint changes, there is a substantial gap between these synthetic datasets and realistic footage.

Our virtual production stage lets us capture moving-camera green-screen takes with tracked camera motion. We record the foreground in-camera and log the camera trajectory; later, we render the background in a game engine to match that motion and composite it with the captured foreground. This approach allows any background with any desired difficulty to be synthesized, improving flexibility and reducing the synthetic-to-real gap.

\begin{table}[]
	\centering
	\small
	\caption{Results on the virtual production dataset CineMatte-4K.}
	\label{result_cinematte}
	\begin{tabular}{@{}lllll@{}}
		\toprule
		& \multicolumn{4}{c}{CineMatte-4K-Image}                                  \\ \midrule
		\multicolumn{1}{l|}{Method}       & MAD$\downarrow$             & MSE$\downarrow$             & Grad$\downarrow$            & Conn$\downarrow$            \\ \midrule
		\multicolumn{1}{l|}{MODNet~\cite{ke2022modnet}}       & 20.828         & 18.609         & 54.108         & 43.506         \\
		\multicolumn{1}{l|}{P3M~\cite{li2021p3m}}          & 11.543         & 10.664         & 10.420         & 82.232         \\
		\multicolumn{1}{l|}{RVM~\cite{lin2022rvm}}          & 8.676          & 5.890          & 8.661          & 55.857         \\
		\multicolumn{1}{l|}{MGM~\cite{yu2021mgmatting}}          & 130.945        & 63.964         & 33.957         & 872.055        \\
		\multicolumn{1}{l|}{MatAnyone~\cite{yang2025matanyone}}    & 1.975          & 0.830          & 8.538          & 15.278         \\
		\multicolumn{1}{l|}{KNN~\cite{chen2013knn}}          & 5.934          & 2.730          & 18.608         & 33.803         \\
		\multicolumn{1}{l|}{ViTMatte~\cite{yao2024vitmatte}}     & 2.797          & 1.461          & 8.409          & 17.399         \\
		\multicolumn{1}{l|}{DIM~\cite{xu2017dim}}          & 2.064          & 0.729          & 6.697          & 11.691         \\
		\multicolumn{1}{l|}{BGMv2~\cite{lin2020bgmv2}}        & 32.973         & 30.585         & 17.027         & 189.237        \\ \midrule
		\multicolumn{1}{l|}{\textbf{CineMatte}}  & \cellcolor[HTML]{C0C0C0}\textbf{0.850} & \cellcolor[HTML]{C0C0C0}\textbf{0.232} & \cellcolor[HTML]{C0C0C0}\textbf{4.013} & \cellcolor[HTML]{C0C0C0}\textbf{3.848} \\ \bottomrule
	\end{tabular}
	\vspace{-0.21cm}
\end{table}
\begin{table}[]
	\centering
	\small
	\caption{Robustness to background shift on YouTubeMatte~\cite{yang2025matanyone}.}
	\label{table_ablation_bg_shift}
	\begin{tabular}{@{}lllll@{}}
		\toprule
		& \multicolumn{2}{c}{Ours}                             & \multicolumn{2}{c}{BGMv2~\cite{lin2020bgmv2}} \\ \midrule
		\multicolumn{1}{l|}{}                                                                                                    & MAD$\downarrow$             & \multicolumn{1}{l|}{MSE$\downarrow$ }            & MAD$\downarrow$          & MSE$\downarrow$          \\ \midrule
		\multicolumn{1}{l|}{No Shift}                                                                                            & \textbf{1.779} & \multicolumn{1}{l|}{\textbf{0.447}} & 2.142       & 0.648       \\ \midrule
		\multicolumn{1}{l|}{\begin{tabular}[c]{@{}l@{}}Angle: (-2, 2)\\ Scale: (0.95, 1.05)\\ Shear: (-0.02, 0.02)\end{tabular}} & \textbf{1.803} & \multicolumn{1}{l|}{\textbf{0.457}} & 3.394       & 1.608       \\ \midrule
		\multicolumn{1}{l|}{\begin{tabular}[c]{@{}l@{}}Angle: (-5, 5)\\ Scale: (0.90, 1.10)\\ Shear: (-0.07, 0.07)\end{tabular}} & \textbf{1.940} & \multicolumn{1}{l|}{\textbf{0.493}} & 37.540      & 34.317      \\ \bottomrule
	\end{tabular}
	\vspace{-0.32cm}
\end{table}

\section{Training}
\subsection{Dataset Preparation}
Our model is trained on a mixture of Composition-1k~\cite{xu2017dim}, AIM500~\cite{li2021aim}, Distinction-646~\cite{qiao2020hattmatting}, and PPM~\cite{ke2022modnet} as foreground sources, and BG-20K~\cite{li2020gfm} as backgrounds. For fair comparison with prior work, we use CineMatte-4K only for evaluation. Standard data augmentations are applied to both foregrounds and backgrounds (similar to~\cite{lin2020bgmv2}), including horizontal flip; color jitter (saturation, hue, brightness, sharpness); and random affine transforms (rotation, scale, shear).

Because synthetic cutouts show noticeable artifacts, the model can learn a shortcut: infer the alpha from foreground appearance alone and ignore the background. To break this shortcut, we use a \textit{distractor strategy}. For each training sample, we insert a few additional foreground subjects into the chosen background. Visually, these distractors look like valid foregrounds, yet they actually belong to the background. This forces the network to consult background cues and scene context to decide what truly belongs to the subject, stabilizing training and improving robustness on real footage.

Finally, we randomly sample $0$--$3$ target foregrounds from the foreground datasets and composite them onto the same background to form the training sample.
\subsection{Training Details}
We train with a learning rate of $1\times 10^{-5}$ for the decoder and FBAM and a lower rate of $1\times 10^{-6}$ for the pretrained feature upsampler to preserve its prior knowledge. The batch size is $1$ per GPU, training resolution is $768 \times 768$. Training runs for approximately $80{,}000$ steps until convergence on two NVIDIA RTX A6000 GPUs, taking around $2$--$3$ days. In the following quantitative comparisons, we use a 2-layer FBAM and a full-attention feature upsampler pretrained with ImageNet~\cite{deng2009imagenet} as in the original paper.

Our loss function follows ViTMatte~\cite{yao2024vitmatte}:
\begin{equation}
	\mathcal{L}_{\mathrm{total}} = \mathcal{L}_{\mathrm{separatel1}} + \mathcal{L}_{\mathrm{lap}} + \mathcal{L}_{\mathrm{gp}}.
\end{equation}
Here $\mathcal{L}_{\mathrm{separatel1}}$ applies separate L1 penalties to the known and unknown regions of a training-time trimap derived from the ground-truth alpha; no trimap is used at inference. $\mathcal{L}_{\mathrm{lap}}$ is the Laplacian loss, and $\mathcal{L}_{\mathrm{gp}}$ is the gradient penalty loss.

\section{Experiments}
In the experiments, we aim to demonstrate our method's generalization ability. Since our method and most compared methods are trained on datasets such as Composition-1k~\cite{xu2017dim}, AIM500~\cite{li2021aim} etc., we evaluate on VideoMatte240K~\cite{lin2020bgmv2} and YouTubeMatte~\cite{yang2025matanyone} for public-dataset comparison. Lastly, we test all methods on our VP dataset, CineMatte-4K. Results are shown in Table~\ref{result_public_dataset} and Table~\ref{result_cinematte}.

For a fair comparison with all the models, we use official weights and suggested parameters that claim the best performance according to their papers or GitHub repositories. Trimaps for trimap-based models are generated from the ground-truth alpha using an fg–bg erosion size of 25 to simulate human annotation. Masks for mask-guided models are generated from the ground-truth alpha with a threshold of 0.95 and then applying a 7×7 elliptical erosion. All methods are evaluated at a resolution of $1024 \times 1024$.

We employ MAD (mean absolute difference), MSE (mean squared error), Grad (spatial gradient)~\cite{rhemann2009grad}, Conn (connectivity)~\cite{rhemann2009grad} and dtSSD~\cite{erofeev2015dtssd} as metrics.

{
	\setlength{\parindent}{0cm}
	\textbf{VideoMatte240K.~\cite{lin2020bgmv2}}
}
VideoMatte240K consists of five test clips totaling 2,715 frames; we randomly sample five BG-20K~\cite{li2020gfm} images as backgrounds.

{
	\setlength{\parindent}{0cm}
	\textbf{YouTubeMatte.~\cite{yang2025matanyone}}
}
For time efficiency and increased difficulty, we evenly sampled 640 frames in total across YouTubeMatte-Static and -Motion, then synthesized with random BG-20K~\cite{li2020gfm} test-set backgrounds.

{
	\setlength{\parindent}{0cm}
	\textbf{CineMatte-4K.}
}
The CineMatte-4K test set consists of 100 images; hand–eye calibration yields only approximate alignment, yet substantial background shift persists.

\subsection{Results and Analysis.}

For both VideoMatte240K~\cite{lin2020bgmv2} and YouTubeMatte~\cite{yang2025matanyone}, our method ranks first among the compared methods on MAD, MSE, and Grad, while trailing slightly on Connectivity. On the temporal metric dtSSD, we place second despite having no explicit temporal module. Our results also significantly surpass the background‑matting baseline BGMv2. We hypothesize that the relatively lower Connectivity compared with trimap‑based methods arises because a trimap provides a stronger semantic prior (hard foreground/background labels), whereas background‑based methods must infer foreground–background semantics and can occasionally produce small floating fragments classified as foreground.

On CineMatte‑4K, our method surpasses all baselines by a significant margin. BGMv2 collapses under large background shifts, leading to an inferior overall score.

 We conducted a dedicated robustness experiment to quantify the impact of background shift on YouTubeMatte~\cite{yang2025matanyone}(Table~\ref{table_ablation_bg_shift}). The analysis confirms that BGMv2's accuracy deteriorates severely as shift magnitude grows, while our method maintains strong performance and a consistent advantage at larger shifts.

\begin{figure}[t]
	\centering
	\includegraphics[width=1\linewidth]{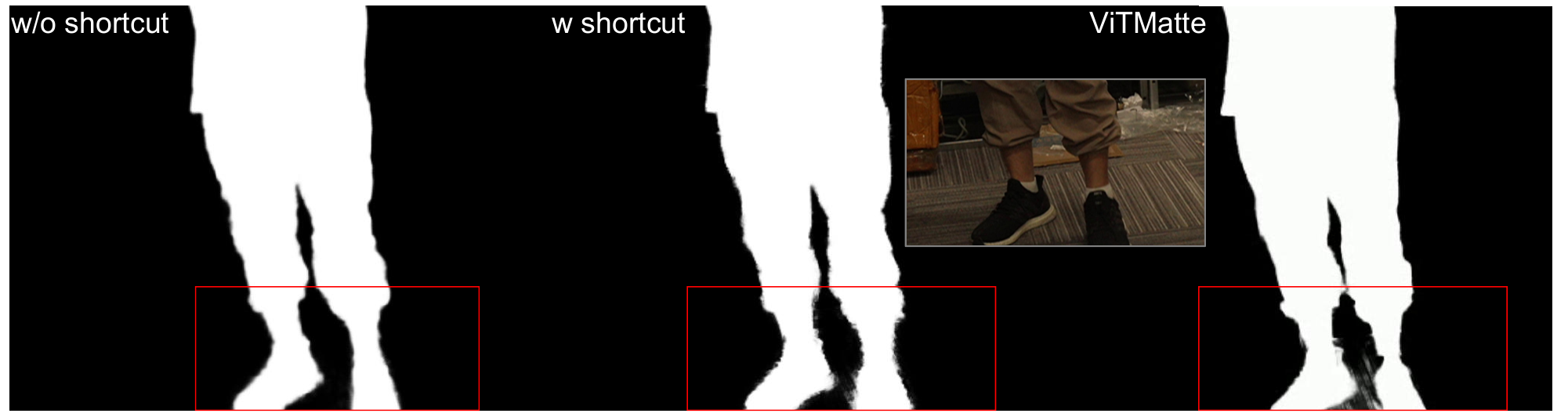}

	\caption{The effect of connecting a high-resolution shortcut from a simple convolutional encoder: boundary details display noticeable artifacts influenced by nearby textures.}
	\label{fig:ablation_shortcut}
	\vspace{-0.3cm}
\end{figure}

\section{Ablation}
{
	\setlength{\parindent}{0cm}
	\textbf{Number of FBAM layers.}
}
We study the effect of FBAM depth (Table~\ref{ablation_fbam}). Increasing depth could model richer foreground–background interactions but also increases overfitting risk. Two FBAM layers seemed optimal.

{
	\setlength{\parindent}{0cm}
	\textbf{Ablation study on different modules.}
}
In Table~\ref{ablation_module} we show the ablation study on different modules. \textit{Baseline} refers to a simple DINOv3 ViT with decoder. \textit{Conv} refers to adding a simple convolutional encoder to restore high-res detail. \textit{Concat} concatenates the background frame with the input and fine‑tunes the entire model end‑to‑end.

{
	\setlength{\parindent}{0cm}
	\textbf{Effect of high-res shortcut.}
}
We assess a ViTMatte‑style~\cite{yao2024vitmatte} convolutional shortcut by adding a high‑resolution conv path to the final layer. Metrics remain similar, but the shortcut yields visible boundary artifacts consistent with feature misalignment (see Fig.~\ref{fig:ablation_shortcut}).

{
	\setlength{\parindent}{0cm}
	\textbf{Effects of different backbones.}
}
We assessed performance under multiple backbone choices, as shown in Table~\ref{ablation_backbone}. Our framework can be readily adapted to future backbones.

\begin{table}[]
	\centering
	\caption{Ablation study for modules on YouTubeMatte~\cite{yang2025matanyone}.}
	\label{ablation_module}
	\begin{tabular}{@{}lll@{}}
		\toprule
		& MAD$\downarrow$             & MSE$\downarrow$             \\ \midrule
		\multicolumn{1}{l|}{Baseline}                      & 5.326          & 2.340          \\
		\multicolumn{1}{l|}{+ Conv}                        & 5.229          & 2.026          \\
		\multicolumn{1}{l|}{+ Feature Upsampler}           & 2.456          & 0.928          \\
		\multicolumn{1}{l|}{+ Feature Upsampler \& Concat} & 6.239          & 3.692          \\
		\multicolumn{1}{l|}{\textbf{+ Feature Upsampler \& FBAM}}   & \textbf{1.779} & \textbf{0.447} \\ \bottomrule
	\end{tabular}
\end{table}

\begin{table}[]
	\centering
	\caption{Ablation study for backbone on VideoMatte240K~\cite{lin2020bgmv2}.}
	\label{ablation_backbone}
	\begin{tabular}{@{}llll@{}}
		\toprule
		& MAD$\downarrow$           & MSE  $\downarrow$           & Grad$\downarrow$            \\ \midrule
		\multicolumn{1}{l|}{SAM2~\cite{ravi2024sam}} & 9.053         & 2.034          & 12.158         \\
		\multicolumn{1}{l|}{DINOv2~\cite{oquab2023dinov2} ViT-B}            & 6.273          & 1.201          & 7.610          \\
		\multicolumn{1}{l|}{DINOv3~\cite{simeoni2025dinov3} ViT-B}            & \textbf{3.817}          & 0.529          & 6.923          \\
		\multicolumn{1}{l|}{\textbf{DINOv3~\cite{simeoni2025dinov3} ViT-L}}            & 4.397 & \textbf{0.341} & \textbf{5.774} \\ \bottomrule
	\end{tabular}
	\vspace{-0.2cm}
\end{table}
\section{Conclusion}
We present CineMatte, a cross-attention‑conditioned background-matting framework that keeps a frozen, weight‑shared DINOv3 ViT for the input and captured background, and conditions the image stream via a cross‑attention Foreground–Background Alignment Module (FBAM); a pretrained, image‑guided upsampler restores crisp, boundary‑aligned detail. We also collect CineMatte‑4K, a 4K HDR VP image–video dataset with non‑synthetic mattes and tracked camera trajectories. Across VideoMatte240K, YouTubeMatte, and CineMatte‑4K, CineMatte achieves competitive scores and shows strong robustness to background misalignment.

	{
		\small
		\bibliographystyle{ieeenat_fullname}
		\bibliography{main}
	}

\end{document}